\documentclass[10pt,twocolumn,letterpaper]{article}

\usepackage{cvpr}
\usepackage{times}
\usepackage{epsfig}
\usepackage{graphicx}
\usepackage{amsmath}
\usepackage{amssymb}

\usepackage{amsfonts}
\usepackage{booktabs}
\usepackage{siunitx}
\usepackage{multirow}
\usepackage{tabularx}
    \newcolumntype{L}{>{\raggedright\arraybackslash}X}
\usepackage{bbold}

\usepackage[pagebackref=true,breaklinks=true,letterpaper=true,colorlinks,bookmarks=false]{hyperref}

\cvprfinalcopy 

\def\cvprPaperID{1221} 

\ifcvprfinal\fi
\begin{document}

\title{FDA: Fourier Domain Adaptation for Semantic Segmentation}

\author{Yanchao Yang\\
UCLA Vision Lab\\
{\tt\small yanchao.yang@cs.ucla.edu}
\and
Stefano Soatto\\
UCLA Vision Lab\\
{\tt\small soatto@cs.ucla.edu}
}

\maketitle
\thispagestyle{empty}

\begin{figure*}[ht!]
\begin{center}
  \includegraphics[width=0.85\textwidth]{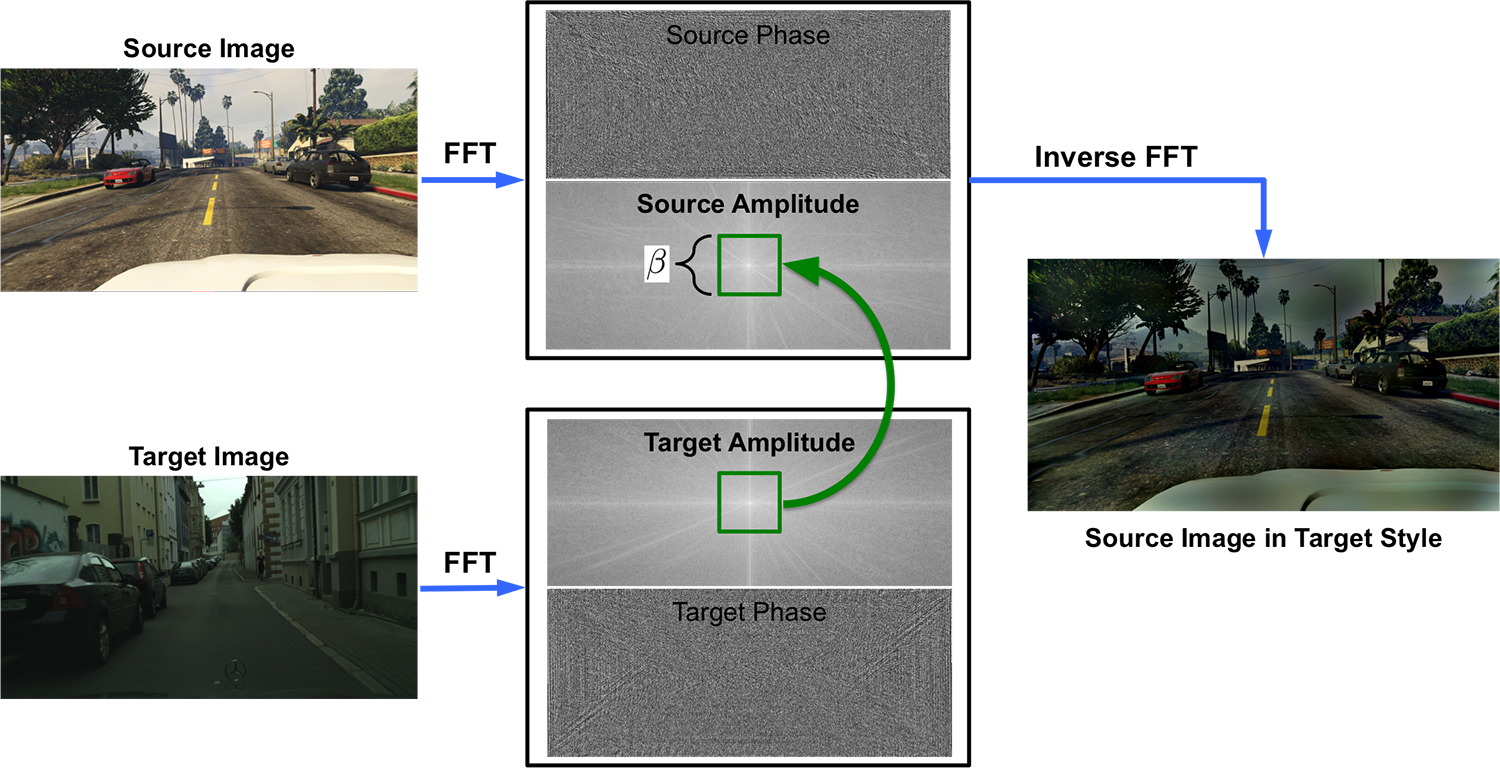}
\end{center}
\caption{{\it Spectral Transfer:} Mapping a source image to a target ``style'' without altering semantic content. A randomly sampled target image provides the style by swapping the low-frequency component of the spectrum of the source image with its own. The outcome ``source image in target style'' shows a smaller domain gap perceptually and improves transfer learning for semantic segmentation as measured in the benchmarks in Sect. \ref{sect:experiments}.}
\vspace{-0.2cm}
\label{fig:illustrate-fftda}
\end{figure*}

\begin{abstract}
We describe a simple method for unsupervised domain adaptation, whereby the discrepancy between the source and target distributions is reduced by swapping the low-frequency spectrum of one with the other. 
We illustrate the method in semantic segmentation, where densely annotated images are aplenty in one domain (\eg, synthetic data), but difficult to obtain in another (\eg, real images). 
Current state-of-the-art methods are complex, some requiring adversarial optimization to render the backbone of a neural network invariant to the discrete domain selection variable. 
Our method does not require any training to perform the domain alignment, just a simple Fourier Transform and its inverse. 
Despite its simplicity, it achieves state-of-the-art performance in the current benchmarks, when integrated into a relatively standard semantic segmentation model. 
Our results indicate that even simple procedures can discount nuisance variability in the data that more sophisticated methods struggle to learn away.\footnote{Code available at: \url{https://github.com/YanchaoYang/FDA}}
\end{abstract}


\section{Introduction}

Unsupervised domain adaptation (UDA) refers to adapting a model trained with annotated samples from one distribution (source), to operate on a different (target) distribution for which no annotations are given. 
For example, the source domain can consist of synthetic images and their corresponding pixel-level labels (semantic segmentation), and the target can be real images with no ground-truth annotations. 
Simply training the model on the source data does not yield satisfactory performance on the target data, due to the covariate shift. 
In some cases, perceptually insignificant changes in the low-level statistics can cause significant deterioration of the performance of the trained model, unless UDA is performed.

State-of-the-art UDA methods train a deep neural network (DNN) model for a given task (say, semantic segmentation) plus an auxiliary loss designed to make the model invariant to the binary selection of source/target domain. 
This requires difficult adversarial training. We explore the hypothesis that simple alignment of the low-level statistics between the source and target distributions can improve performance in UDA, without any need for training beyond the primary task of semantic segmentation.

Our method is illustrated in Fig.~\ref{fig:illustrate-fftda}: One simply computes the (Fast) Fourier Transform (FFT) of each input image, and replaces the low-level frequencies of the target images into the source images before reconstituting the image for training, via the inverse FFT (iFFT), using the original annotations in the source domain.

To test our hypothesis, we use as a baseline (lower bound) the performance on the target data of a model trained on the source. 
As a paragon (upper bound), we use a state-of-the-art model with adversarial training \cite{hoffman2018cycada}. We expect that such a simple, ``zero-shot'' alignment of low-level statistics would improve the baseline, and hopefully come close to the paragon. However, the method actually outperforms the paragon in semantic segmentation. We do not take this to mean that our method is {\em the way} to perform UDA, in particular for general tasks beyond semantic segmentation. However, the fact that such a simple method outperforms sophisticated adversarial learning suggests that these models are not effective at managing low-level nuisance variability.

Fourier domain adaptation requires selecting one free parameter, the size of the spectral neighborhood to be swapped (green square in Fig.~\ref{fig:illustrate-fftda}). We test a variety of sizes, as well as a simple multi-scale method consisting of averaging the results arising from different domain sizes. 

The motivation for our approach stems from the observation that the low-level spectrum (amplitude) can vary significantly without affecting the perception of high-level semantics. 
Whether something is a vehicle or a person should not depend on the characteristics of the sensor, or the illuminant, or other low-level sources of variability. Yet such variability has significant impact on the spectrum, forcing a learning-based model to ``learn it away'' along with other nuisance variability. 
If this variability is not represented in the training set, the models fail to generalize. However, there are sources of variability that we know at the outset not to be informative of the task at hand. 
The categorical interpretation of an image is unchanged if we manipulated global photometric statistics. Any monotonic rescaling of the color map, including non-linear contrast changes, are known nuisance factors, and can be eliminated at the outset without having to be learned. This is especially important since it appears that networks do not transfer well across different low-level statistics \cite{achille2017critical}. While one could normalize contrast transformations, in the absence of a canonical reference our Fourier transfer is among the simplest methods to register them. The broader point is that known nuisance variability can be dealt with at the outset, without the need to learn it through complex adversarial training.

In the next section, we describe our method in more detail and then test it empirically in standard UDA benchmarks. Before doing so, we place our work in the context of the current literature.

\subsection{Related Work}

{\it Semantic Segmentation} has benefited by the continuous evolution of DNN architectures \cite{long2015fully,yu2015multi,chen2017deeplab,zhao2017pyramid,sun2019deep}. These are generally trained on datasets with dense pixel-level annotations, such as Cityscapes \cite{cordts2016cityscapes}, PASCAL \cite{everingham2015pascal} and MSCOCO \cite{lin2014microsoft}. Manual annotation is not scalable \cite{zhang2017curriculum}, and capturing representative imaging conditions adds to the challenges. This has spurred interest in 
using synthetic data, such as from GTA5 \cite{richter2016playing} and SYNTHIA \cite{ros2016synthia}. Due to the domain shift, models trained on the former tend to perform poorly on the latter. 

{\it Domain Adaptation} aims to reduce the shift between two distributions \cite{patel2015visual,csurka2017domain,wang2018deep}. A common discrepancy measure is MMD (Maximum Mean Discrepancy) and its kernel variants \cite{geng2011daml,long2015learning}, extended by CMD (Central Moment Discrepancy) \cite{zellinger2017central} to higher-order statistics \cite{cariucci2017autodial,mancini2018boosting}. Unfortunately, two datasets are not guaranteed to be aligned even if the MMD is minimized, due to the limited expressiveness of such metrics. %
{\it Adversarial Learning} for domain adaptation \cite{ganin2014unsupervised,tzeng2017adversarial,shu2018dirt,kumar2018co} uses a discriminator trained to maximize the confusion between source and target representations, thus reducing the domain discrepancy. Alignment in high-level feature space \cite{long2015learning,ghifary2016deep,saito2017asymmetric,sener2016learning,motiian2017unified} can be counter-productive for semantic segmentation, unlike image-level classification,   \cite{hoffman2016fcns,luo2019significance,sankaranarayanan2018learning}, due to the complex representations and the difficulty in stabilizing adversarial training. 

We draw on image-to-image translation and style transfer \cite{zhu2017unpaired,liu2017unsupervised,yi2017dualgan,choi2018stargan} to improve {\it domain adaptation for semantic segmentation.} Cycada \cite{hoffman2018cycada} aligns representations at both the pixel-level and feature-level. DCAN \cite{wu2018dcan} preserves spatial structures and semantics by the channel-wise alignment of multi-level features. To facilitate image space alignment, \cite{chang2019all} proposes domain-invariant structure extraction to disentangle domain-invariant and domain-specific representations. 
\cite{chen2019learning} uses dense depth, readily available in synthetic data. \cite{gong2019dlow} generates intermediate style images between source and target. CLAN \cite{luo2019taking} enforce local semantic consistency in global alignment. \cite{zhang2017curriculum} proposes curriculum-style learning to align both global distributions over images and local distributions over landmark superpixels. BDL \cite{li2019bidirectional} employs bidirectional learning, where the segmentation network is exploited by the image transformation network. There are also discriminators applied on the output space \cite{chen2019learning,tsai2018learning} to align source and target segmentation.  

The use of a transformer network and discriminators at multiple levels is computationally demanding, and more challenging to train within the adversarial framework. In contrast, our method does not utilize any image translation networks to generate training images, nor discriminators to align pixel/feature-level distributions. The only network trained in our method is for the primary task of semantic segmentation. We use a fully convolutional network that outputs pixel-wise class (log) likelihoods.
Note in the concurrent work \cite{yang2020phase}, a transformer network trained with phase preservation as a constraint also generates domain aligned images that maintain semantic content in the source images.
Similar adaptation gain is then achieved by enforcing the scene compatibility learned using Conditional Prior Networks \cite{yang2018conditional}.

Domain adaptation and {\it Semi-Supervised Learning} (SSL) are closely related. When the domains are aligned, unsupervised domain adaptation becomes SSL. CBST \cite{zou2018unsupervised} and BDL \cite{li2019bidirectional} used ``self-training'' as a form of regularization \cite{rosenberg2005semi}, exploiting target images by treating pseudo-labels as ground truth. ADVENT \cite{vu2019advent}  minimizes both the entropy of the pixel-wise predictions and the adversarial loss of the entropy maps. The computation of the pixel-wise entropy does not depend on any networks and entails no overhead. We employ entropy minimization to regularize the training of our segmentation network. Motivated by \cite{tarvainen2017mean,laine2016temporal,french2017self}, we also average the output of different models that are trained with different spectral domain size, which fosters multi-band transfer as discussed in detail next.

\begin{figure*}[!ht]
\begin{center}
  \includegraphics[width=0.9\textwidth]{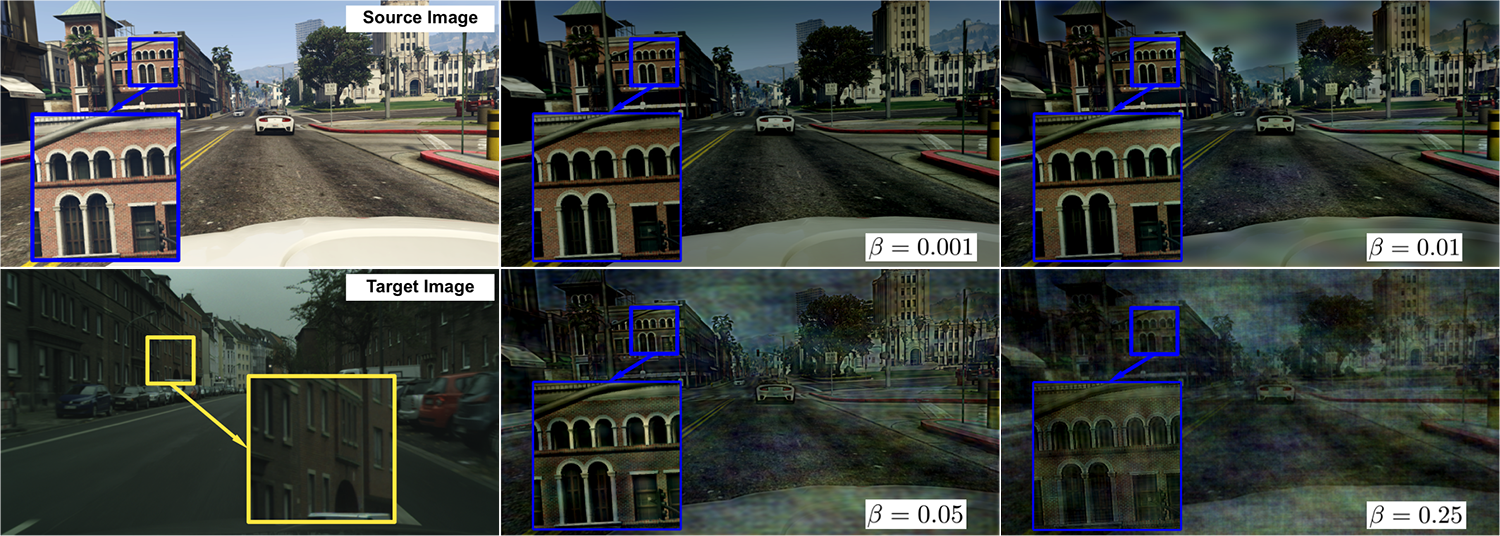}
\end{center}
\vspace{-0.2cm}
\caption{ {\it Effect of the size of the domain $\beta$, shown in Fig. \ref{fig:illustrate-fftda}, where the spectrum is swapped}: increasing $\beta$ will decrease the domain gap but introduce artifacts (see zoomed insets). We tune $\beta$ until artifacts in the transformed images become obvious and use a single value for some experiments. In other experiments, we maintain multiple values simultaneously in a multi-scale setting (Table \ref{tab:abalation-study-FDA}).}
\vspace{-0.2cm}
\label{fig:effect_beta}
\end{figure*}

\section{Method}

We first describe the simple Fourier alignment, which does not require any training, and then describe the loss we use to train the overall semantic segmentation network to leverage the Fourier alignment. 

\subsection{Fourier Domain Adaptation (FDA)}

In unsupervised domain adaptation (UDA), we are given a source dataset $D^s=\{ (x^s_i, y^s_i)  \sim P(x^s, y^s)\}_{i=1}^{N_s}$, where $x^s \in \mathbb{R}^{H \times W \times 3}$ is a color image, and $y^s \in \mathbb{R}^{H \times W}$ is the semantic map associated with $x^s$. Similarly $D^t=\{ x^t_i \}_{i=1}^{N_t}$ is the target dataset, where the ground truth semantic labels are absent. Generally, the segmentation network trained on $D^s$ will have a performance drop when tested on $D^t$. Here, we propose Fourier Domain Adaptation (FDA) to reduce the domain gap between the two datasets.

Let $\mathcal{F}^{A}, \mathcal{F}^{P}: \mathbb{R}^{H \times W \times 3} \rightarrow \mathbb{R}^{H \times W \times 3}$ be the amplitude and phase components of the Fourier transform $\mathcal{F}$ of an RGB image, \ie, for a single channel image $x$ we have:
\begin{equation}
    \mathcal{F}(x)(m, n) = \sum_{h,w} x(h,w) e^{-j2\pi\left(\dfrac{h}{H}m + \dfrac{w}{W}n\right)}, j^2=-1
\end{equation}
which can be implemented efficiently using the FFT algorithm in \cite{frigo1998fftw}. Accordingly, $\mathcal{F}^{-1}$ is the inverse Fourier transform that maps spectral signals (phase and amplitude) back to image space. Further, we denote with $M_{\beta}$ a mask, whose value is zero except for the center region where $\beta \in (0, 1)$:
\begin{equation}
    M_{\beta}(h,w) = \mathbb{1}_{(h,w) \in [-\beta H: \beta H, -\beta W : \beta W]}
\end{equation}
here we assume the center of the image is $(0,0)$. Note that $\beta$ is not measured in pixels, thus the choice of $\beta$ does not depend on image size or resolution. Given two randomly sampled images $x^s \sim D^s, x^t \sim D^t$, Fourier Domain Adaptation can be formalized as:
\begin{equation}
    x^{s\to t}=\mathcal{F}^{-1}([M_{\beta} \circ \mathcal{F}^A(x^t) + (1-M_{\beta}) \circ \mathcal{F}^A(x^s), \mathcal{F}^P(x^s)])
\label{eq:fda}
\end{equation}
where the low frequency part of the amplitude of the source image $\mathcal{F}^A(x^s)$ is replaced by that of the target image $x^t$. Then, the modified spectral representation of $x^s$, with its phase component unaltered, is mapped back to the image $x^{s\to t}$, whose content is the same as $x^s$, but will resemble the appearance of a sample from $D^t$. The process is illustrated in Fig. \ref{fig:illustrate-fftda} where the mask $M_\beta$ is shown in green.

{\bf Choice of $\beta$}: As we can see from Eq. \eqref{eq:fda}, $\beta=0$ will render $x^{s\to t}$ the same as the original source image $x^s$. On the other hand, when $\beta=1.0$, the amplitude of $x^s$ will be replaced by that of $x^t$. Fig. \ref{fig:effect_beta} illustrates the effect of $\beta$. We find that, as $\beta$ increases to $1.0$, the image $x^{s\to t}$ approaches the target image $x^t$, but also exhibits visible artifacts, as can be seen from the enlarged area in Fig. \ref{fig:effect_beta}. We set $\beta \leq 0.15$. However, in Table \ref{tab:abalation-study-FDA} we show the effect of various choices of $\beta$ along with the average of the resulting models, akin to a simple multi-scale pooling method.

\subsection{FDA for Semantic Segmentation}

\begin{figure}
\begin{center}
  \includegraphics[width=0.28\textwidth]{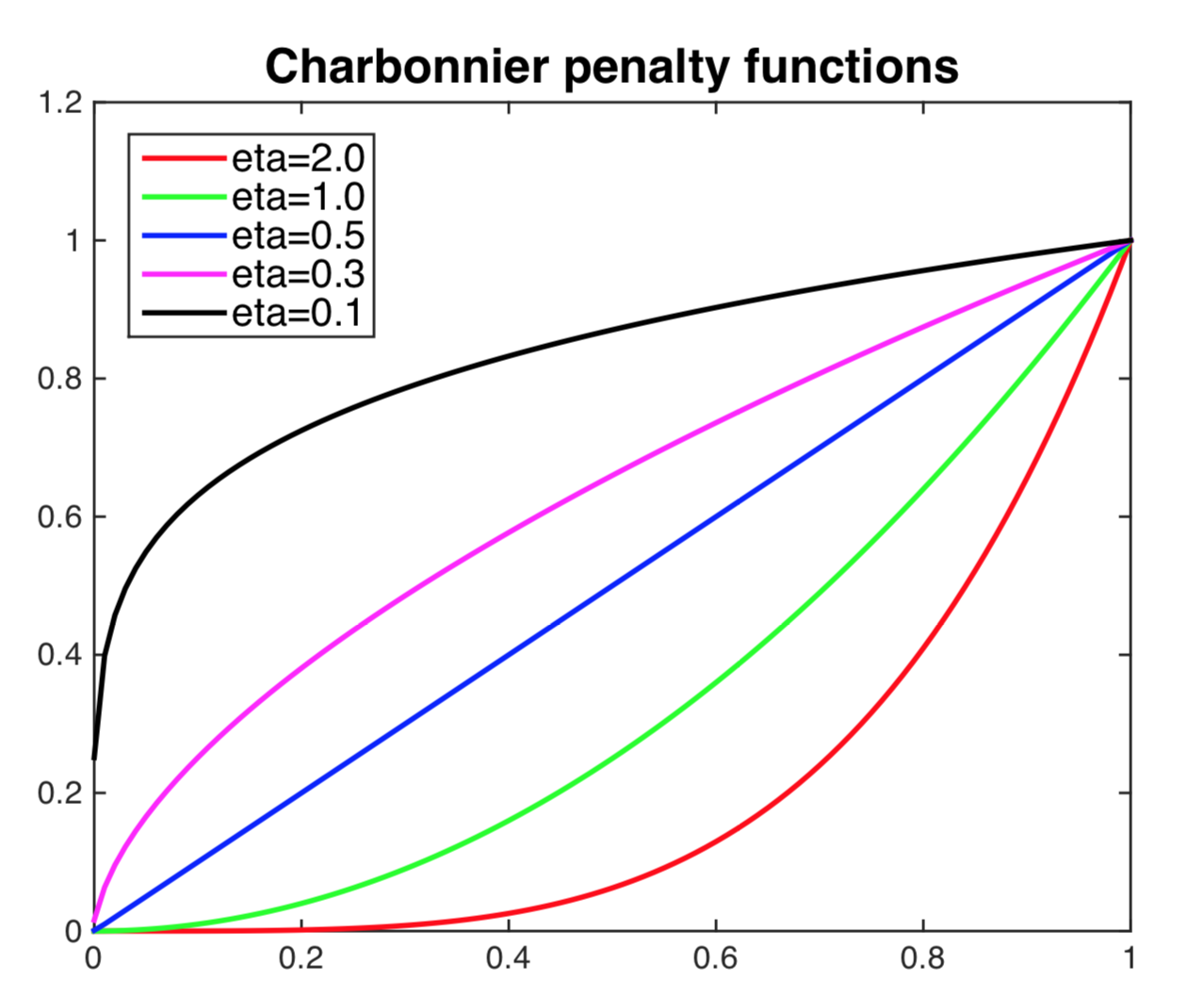}
\end{center}
\vspace{-0.1cm}
\caption{Charbonnier penalty used for robust entropy minimization, visualized for different values of the parameter $\eta$. }
\vspace{-0.1cm}
\label{fig:cbnorm}
\end{figure}

Given the adapted source dataset $D^{s\to t}$,\footnote{the cardinality of $D^{s\to t}$ should be $|D^s| \times |D^t|$, which is large, so we do online random generation of $D^{s\to t}$ given the efficiency of the FFT.} we can train a semantic segmentation network $\phi^{w}$, with parameters $w$, by minimizing the following cross-entropy loss:
\begin{equation}
    \mathcal{L}_{ce}(\phi^w; D^{s\to t}) = -\sum_i \langle y^s_i, \log( \phi^w(x^{s\to t}_i) )\rangle.
\label{eq:cross-entropy}
\end{equation}
Since FDA aligns the two domains, UDA becomes a semi-supervised learning (SSL) problem. The key to SSL is the regularization model. We use as a criterion a penalty for the decision boundary to cross clusters in the unlabeled space. This can be achieved, assuming class separation, by penalizing the decision boundary traversing regions densely populated by data points, which can be done by minimizing the prediction entropy on the target images. However, as noted in \cite{vu2019advent}, this is ineffective in regions with low entropy. Instead of placing an arbitrary threshold on which pixels to apply the penalty to, we use a robust weighting function for entropy minimization, namely
\begin{equation}
    \mathcal{L}_{ent}(\phi^w; D^{t}) = \sum_i \rho( -\langle \phi^w(x^{t}_i), \log( \phi^w(x^{t}_i) )\rangle )
\label{eq:robust-entropy}
\end{equation}
where $\rho(x) = (x^2 + 0.001^2)^{\eta}$ is the Charbonnier penalty function \cite{bruhn2005towards}. It penalizes high entropy predictions more than the low entropy ones for $\eta > 0.5$ as shown in Fig. \ref{fig:cbnorm}. Combining this with the segmentation loss on the adapted source images, we can use the following overall loss to train the semantic segmentation network $\phi^w$ from scratch:
\begin{equation}
    \mathcal{L}(\phi^w; D^{s\to t}, D^t) = \mathcal{L}_{ce}(\phi^w; D^{s\to t}) + \lambda_{ent} \mathcal{L}_{ent}(\phi^w; D^{t})
\label{eq:training-loss-seg-ent}
\end{equation}

{\bf Self-Supervised training} (or, more accurately, ``self-learning'') is a common way of attempting to boost the performance of SSL by using highly confident pseudo-labels predicted with $\phi^w$ as if they were ground truth. In the absence of regularization, this practice is self-referential, so we focus on regularization.

As observed in \cite{tarvainen2017mean}, the mean teacher improves semi-supervised learning performance by averaging the model weights, which provides regularization in the learning process. Here, we propose using the mean of the predictions of multiple models to regularize self-learning. However, instead of training multiple models using the same loss at once, with an explicit divergence term as in \cite{kumar2018co}, we directly train multiple models $\phi^{w}_{\beta}$ with different $\beta's$ in the FDA process, with no need to explicitly force model divergence. We instantiate M=3 segmentation networks $\phi^w_{\beta_m}, m=1,2,3$, which are all trained from scratch using \eqref{eq:training-loss-seg-ent}, and the mean prediction for a certain target image $x^t_{i}$ can be obtained by:
\begin{equation}
    \hat{y}^t_{i} = \arg\max_{k} \dfrac{1}{M} \sum_m \phi^w_{\beta_m}(x^t_{i}).
\end{equation}
Note that the output of the network is the softmax activation, so the average is still a probability distribution over K categories. Using the pseudo-labels generated by M models, we can train $\phi^w_{\beta}$ to get further improvement using the following self-supervised training loss:
\begin{multline}
    \mathcal{L}_{sst}(\phi^w; D^{s\to t}, \hat{D}^t) =  \mathcal{L}_{ce}(\phi^w; D^{s\to t}) \\
    + \lambda_{ent} \mathcal{L}_{ent}(\phi^w; D^{t}) + \mathcal{L}_{ce}(\phi^w; \hat{D}^{t})
\label{eq:self-supervised-training}
\end{multline}
where $\hat{D}^t$ is $D^t$ augmented with pseudo labels $\hat{y}^t_i$'s. Since our training entails different $\beta$'s in the FDA operation, we call the self-supervised training using the mean prediction of different segmentation networks Multi-band Transfer (MBT). The full training procedure of our FDA semantic segmentation network consists of one round of initial training of M models from scratch using Eq. \eqref{eq:training-loss-seg-ent}, and two more rounds of self-supervised training using Eq. \eqref{eq:self-supervised-training}, as we detail in the next section.

\section{Experiments}
\label{sect:experiments}

\subsection{Datasets and Training Details}

We evaluate the proposed method on two challenging synthetic-to-real unsupervised domain adaptation tasks, where we have abundant semantic segmentation labels in the synthetic domain (source), but zero in the real domain (target). The two synthetic datasets are GTA5 \cite{richter2016playing} and SYNTHIA \cite{ros2016synthia}; the real domain dataset is CityScapes \cite{cordts2016cityscapes}.

{\bf GTA5}: consists of 24,966 synthesized images captured in a video game, with the original image size 1914$\times$1052. During training, we resize the images to 1280$\times$720, and then random crop them to 1024$\times$512. The original GTA5 provides pixel-wise semantic annotations of 33 classes, but we use the 19 classes in common with CityScapes for standard comparison to other state-of-the-art methods.

{\bf SYNTHIA}: also aligned with the other SOTA methods, we use the SYNTHIA-RAND-CITYSCAPES subset which has 9,400 annotated images with the original resolution 1280$\times$760. The images are randomly cropped to 1024$\times$512 during training. Again, the 16 common classes are used for training, but evaluations are performed on both the 16 classes and a subset of 13 classes following the standard protocol.

{\bf CityScapes}: is a real-world semantic segmentation dataset collected in driving scenarios. We use the 2,975 images from the training set as the target domain data for training. We test on the 500 validation images with dense manual annotations. Images in CityScapes are simply resized to 1024$\times$512, with no random cropping. The two domain adaptation scenarios are GTA5$\rightarrow$CityScapes and SYNTHIA$\rightarrow$CityScapes.

Note that, in all experiments, we perform FDA via Eq. \eqref{eq:fda} on the training images in the range $[0,255]$ before we do mean subtraction, since the FFT algorithm we employ is numerically stable for non-negative values.

{\bf Segmentation Network $\phi^w$}: We experiment with two different architectures to show the robustness of FDA, DeepLabV2 \cite{chen2017deeplab} with a ResNet101 \cite{he2016deep} backbone, and FCN-8s \cite{long2015fully} with a VGG16 \cite{simonyan2014very} backbone. We use the same initialization as in \cite{li2019bidirectional} for both networks. Again, the segmentation network $\phi^w$ is the only network in our method.

{\bf Training}: Our training is carried out on a GTX1080 Ti GPU; due to memory limitations, the batch size is set to 1 in all our experiments. To train DeepLabV2 with ResNet101 using SGD, the initial learning rate is 2.5e-4, and adjusted according to the 'poly' learning rate scheduler with a power of 0.9, and weight decay 0.0005. For FCN-8s with VGG16, we use ADAM with the initial learning rate 1e-5, which is decreased by the factor of 0.1 every 50000 steps until 150000 steps. We also apply early stopping as in \cite{li2019bidirectional}. The momentum for Adam is 0.9 and 0.99.

\subsection{FDA with Single Scale}

We first test the proposed FDA method with single scale on the task GTA5$\rightarrow$CityScapes. We instantiate three DeepLabV2 segmentation netowrks $\phi^w_{\beta}$, with $\beta=0.01, 0.05, 0.09$, and train them separately using Eq. \eqref{eq:training-loss-seg-ent}. We set $\lambda_{ent}=0.005$ and $\eta=2.0$ for all experiments. We report the mean intersection over union score (mIOU) across semantic classes on the validation set of CityScapes in Tab. \ref{tab:abalation-study-FDA}, where T=0 represents training from scratch. As we can see from the first section in Tab. \ref{tab:abalation-study-FDA}, The segmentation networks trained with different $\beta$'s in the FDA operation maintain similar performance. This demonstrates the robustness of FDA with respect to the choice of $\beta$ when training with Eq. \eqref{eq:training-loss-seg-ent}.

Moreover, the network $\phi^w_{\beta=0.09}$ trained simply using Eq. \eqref{eq:cross-entropy} ($\beta=0.09$, $\lambda_{ent}=0$) i.e., without entropy loss, surpasses the baseline Cycada \cite{hoffman2018cycada} by 4.54\%, which demonstrates better management of variability by FDA than the two-stage image translation based adversarial domain adaptation, where an image transformer is trained from one domain to another, and a discriminator is trained to distinguish between the two domains.

\setlength{\tabcolsep}{0.18em}
\begin{table*}
\begin{center}
\resizebox{0.9\textwidth}{!}{
\begin{tabular}{ccccccccccccccccccccc} \toprule
    {Experiment} & {\rotatebox[origin=l]{70}{road}} 
                 & {\rotatebox[origin=l]{70}{sidewalk}} 
                 & {\rotatebox[origin=l]{70}{building}} 
                 & {\rotatebox[origin=l]{70}{wall}} 
                 & {\rotatebox[origin=l]{70}{fence}} 
                 & {\rotatebox[origin=l]{70}{pole}} 
                 & {\rotatebox[origin=l]{70}{light}} 
                 & {\rotatebox[origin=l]{70}{sign}} 
                 & {\rotatebox[origin=l]{70}{vegetation}} 
                 & {\rotatebox[origin=l]{70}{terrain}} 
                 & {\rotatebox[origin=l]{70}{sky}} 
                 & {\rotatebox[origin=l]{70}{person}} 
                 & {\rotatebox[origin=l]{70}{rider}} 
                 & {\rotatebox[origin=l]{70}{car}} 
                 & {\rotatebox[origin=l]{70}{truck}} 
                 & {\rotatebox[origin=l]{70}{bus}} 
                 & {\rotatebox[origin=l]{70}{train}} 
                 & {\rotatebox[origin=l]{70}{motocycle}} 
                 & {\rotatebox[origin=l]{70}{bicycle}} 
                 & {mIoU} \\ \midrule
     {$\beta$=0.01 (T=0)} & 88.8 & 35.4 & 80.5 & 24.0 & 24.9 & 31.3 & \underline{34.9} & 32.0 & \underline{82.6} & \underline{35.6} & 74.4 & \underline{59.4} & \underline{31.0} & 81.7 & \underline{29.3} & \underline{47.1} & 1.2 & 21.1 & 32.3 & 44.61 \\
     {$\beta$=0.05 (T=0)} & 90.7 & \underline{45.0} & 80.4 & 24.6 & 22.6 & \underline{31.8} & 30.3 & \underline{39.4} & 81.4 & 33.8 & 72.6 & 57.6 & 29.1 & \underline{83.2} & 26.3 & 36.9 & 6.6 & 20.6 & \underline{34.9} & 44.6 \\
     {$\beta$=0.09 (T=0)} & \underline{90.8} & 42.7 & \underline{80.8} & \underline{28.1} & \underline{26.6} & 31.8 & 32.8 & 29.1 & 81.6 & 31.2 & \underline{76.2} & 56.9 & 27.7 & 82.8 & 25.3 & 44.1 & \underline{15.3} & \underline{21.1} & 30.2 & 45.01 \\ \midrule
     Cycada\cite{hoffman2018cycada} & 86.7 & 35.6 & 80.1 & 19.8 & 17.5 & 38.0 & 39.9 & 41.5 & 82.7 & 27.9 & 73.6 & 64.9 & 19 & 65.0 & 12.0 & 28.6 & 4.5 & 31.1 & 42.0 & 42.7 \\
     {$\beta$=0.09 ($\lambda_{ent}=0$)} & 90.0 & 40.5 & 79.4 & 25.3 & 26.7 & 30.6 & 31.9 & 29.3 & 79.4 & 28.8 & 76.5 & 56.4 & 27.5 & 81.7 & 27.7 & 45.1 & 17.0 & 23.8 & 29.6 & 44.64 \\
     {$\beta$=0.09 (SST)} & 91.6 & 52.4 & 81.2 & 26.8 & 22.7 & 31.6 & 33.3 & 32.6 & 81.1 & 29.2 & 73.8 & 57.2 & 27.1 & 82.5 & 23.8 & 44.4 & 15.4 & 21.9 & 34.7 & 45.42 \\
     {MBT (T=0)} & 91.3 & 44.2 & 82.2 & 32.1 & 29.4 & 32.8 & 35.7 & 30.4 & 83.2 & 35.7 & 76.3 & 59.8 & 31.7 & 84.5 & 29.5 & 46.1 & 6.9 & 23.2 & 33.7 & 46.77 \\ \midrule
     {$\beta$=0.01 (T=1)} & 92.3 & 51.4 & 82.3 & 30.5 & 24.5 & 31.2 & 36.9 & 34.2 & 82.4 & 39.7 & 76.6 & 57.6 & 28.5 & 82.3 & 27.9 & 47.0 & 5.5 & 21.7 & 40.3 & 47.03 \\
     {$\beta$=0.05 (T=1)} & 92.2 & 50.9 & 81.5 & 27.2 & 27.3 & 32.5 & 35.8 & 35.7 & 81.3 & 37.1 & 76.3 & 58.6 & 30.0 & 83.0 & 23.4 & 45.1 & 6.7 & 23.8 & 40.0 & 46.8 \\
     {$\beta$=0.09 (T=1)} & 91.0 & 46.9 & 80.3 & 25.3 & 21.1 & 30.1 & 35.5 & 37.8 & 80.8 & 38.9 & 79.1 & 58.5 & 31.2 & 82.4 & 29.4 & 46.0 & 9.1 & 24.2 & 39.1 & 46.71 \\
     {MBT (T=1)} & 92.5 & 52.0 & 82.4 & 30.3 & 25.6 & 32.4 & 38.3 & 36.6 & 82.5 & 41.0 & 78.6 & 59.4 & 30.6 & 83.7 & 28.4 & 48.3 & 6.4 & 24.0 & 40.8 & 48.14 \\ \midrule
     {$\beta$=0.01 (T=2)} & 92.1 & 51.5 & 82.3 & 26.3 & 26.8 & 32.6 & 36.9 & 39.6 & 81.7 & 40.7 & 78.2 & 57.8 & 29.1 & 82.8 & 36.1 & 49.0 & 13.9 & 24.5 & 43.9 & 48.77 \\
     {$\beta$=0.05 (T=2)} & 91.6 & 49.7 & 81.1 & 25.2 & 22.7 & 31.5 & 35.0 & 35.1 & 80.8 & 38.2 & 77.5 & 58.9 & 31.3 & 83.0 & 26.9 & 50.5 & 20.8 & 26.4 & 42.2 & 47.86 \\
     {$\beta$=0.09 (T=2)} & 91.6 & 50.6 & 81.0 & 24.4 & 26.0 & 32.2 & 35.3 & 36.5 & 81.3 & 33.1 & 74.5 & 57.8 & 31.2 & 82.9 & 30.0 & 49.7 & 7.0  & 26.1 & 41.6 & 47.03 \\
     {MBT (T=2)} & 92.5 & 53.3 & 82.3 & 26.5 & 27.6 & 36.4 & 40.5 & 38.8 & 82.2 & 39.8 & 78.0 & 62.6 & 34.4 & 84.9 & 34.1 & 53.12 & 16.8 & 27.7 & 46.4 & 50.45 \\ \bottomrule
\end{tabular}}
\end{center}
\vspace{-0.2cm}
\caption{{\it Ablation study on the GTA5$\rightarrow$CityScapes task.} The first section (T=0) shows the performance of the segmentation networks $\phi^w_{\beta}$'s when trained from scratch using Eq. \eqref{eq:training-loss-seg-ent}. Note that as $\beta$ varies, the performance of each $\phi^w_{\beta}$ stays similar, whereas, the best performing entries (underlined) equally distribute among the three individual networks. When the predictions across different $\phi^w_{\beta}$'s are averaged (MBT (T=0)), the mIOU improves over all the constituent ones. And this is true even after the first (T=1) and the second (T=2) round of self-supervised training using Eq. \eqref{eq:self-supervised-training}. Also note that, simply performing self-supervised training without averaging (MBT), the improvement over ($\beta$=0.09 (T=0)) is marginal ($\beta$=0.09 (SST)). }
\vspace{-0.3cm}
\label{tab:abalation-study-FDA}
\end{table*}

\subsection{Multi-band Transfer (MBT)}

We could apply self-training (SST) using the pseudo labels generated for the target domain to further improve the performance of a single network. However, the gain is pretty marginal as expected, as can be seen from the second section in Tab. \ref{tab:abalation-study-FDA}, entry ($\beta$=0.09, SST). The relative improvement after SST is only 0.9\%, compared to ($\beta$=0.09, T=0) in the first section. However, when we analyze the networks trained from scratch with different $\beta$'s in the first section, we can see that, even though the performance is robust to the change of $\beta$, the best performing entries (underlined) are equally distributed across classes, rather than being dominated by a single network. This suggests averaging over predictions of different $\phi^w_{\beta}$'s. By simply averaging prediction from the first round (MBT, T=0), we get a more significant relative improvement of 3.9\% than the best performer from the first round ($\beta=0.09$, T=0). This is consistently observed also in subsequent self-supervised training rounds in the third and fourth sections in Tab. \ref{tab:abalation-study-FDA}.

\subsection{Self-supervised Training with MBT}

We can treat the pseudo labels generated from MBT (T=0) as if they are ground truth labels to train $\phi^w_{\beta}$'s using Eq. \eqref{eq:self-supervised-training}. However, this is self-referential and cannot be expected to work. To regularize, we also apply a thresholding on the confidence values of each prediction. More specifically, for each semantic class, we accept the predictions with confidence that is within the top 66\% or above 0.9. In the third and fourth sections in Tab. \ref{tab:abalation-study-FDA}, we list the performance of each $\phi^w_{\beta}$ after the first round of SST (T=1) and the second round (T=2).

\begin{figure}[!h]
\begin{center}
  \includegraphics[width=0.35\textwidth]{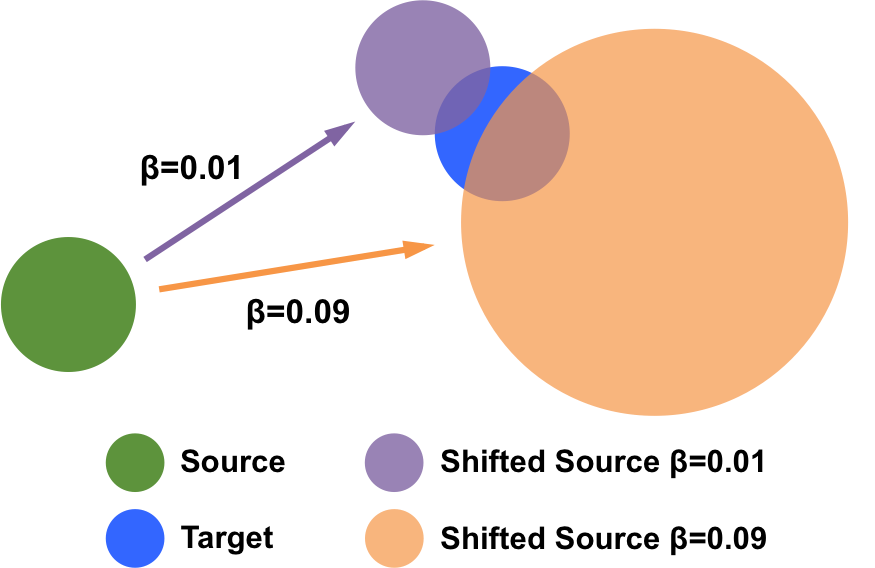}
\end{center}
\caption{ {\it Larger $\beta$ generalizes better if trained from scratch, but induce more bias when combined with Self-supervised Training.}}
\vspace{-0.3cm}
\label{fig:explain-beta-choice}
\end{figure}

However, if we check the relative improvement of each $\phi^w_{\beta}$ by SST (T=0,1,2), we see that the best performer in the training from scratch round (T=0) is $\phi^w_{\beta=0.09}$, which becomes the worst performer during the first SST round (T=1), and finally, after the second round of SST (T=2), $\phi^w_{\beta=0.01}$ becomes the best performer rater than $\phi^w_{\beta=0.09}$. We conjecture that small $\beta$ will yield less variations (artifacts), thus the adapted source dataset $D^{s\to t}$ has less chance to cover the target dataset than the one with larger $\beta$. However, when pseudo labels are used to further align the two domains, $D^{s\to t}$ will impose less bias, since its center is closer to the target dataset and variance is smaller. We illustrate this in Fig. \ref{fig:explain-beta-choice}. Also this observation provides us a reference to set $\beta$, i.e. if we just perform a single scale FDA, we may want to use relatively larger $\beta$, however, for MBT, we may gradually raise the weight on the predictions from $\phi^w_{\beta}$ with smaller $\beta$.

\subsection{Benchmarks}

{\bf GTA5$\rightarrow$CityScapes}: We report the quantitative evaluation of our method in Tab. \ref{tab:benchmark_gta2city}. Again, we can observe that the single scale FDA (FDA) with ResNet101 outperforms most methods that employ adversarial training by instantiating an image transformer and a discriminator \cite{hoffman2018cycada,tsai2018learning,gong2019dlow,luo2019taking}. With entropy minimization activated, the single scale FDA (FDA-ENT) achieves similar performance as \cite{chang2019all,vu2019advent}, which incorporates spatial priors or more sophisticated adversarial training on the structured output of entropy map. By applying SST using the Multi-band Transfer, our method achieves the top performance among others (FDA-MBT). Note that BDL \cite{li2019bidirectional} also performs SST in the adversarial setting, and our method achieves a 4.0\% improvement over BDL. The advantage of our method is also demonstrated on the VGG backbone in the second section of Tab. \ref{tab:benchmark_gta2city}. 

\setlength{\tabcolsep}{0.15em}
\begin{table*}
\begin{center}
\resizebox{0.94\textwidth}{!}{
\begin{tabular}{cccccccccccccccccccccc} \toprule
    {Backbone} & {Method} & {\rotatebox[origin=l]{70}{road}} 
                          & {\rotatebox[origin=l]{70}{sidewalk}} 
                          & {\rotatebox[origin=l]{70}{building}} 
                          & {\rotatebox[origin=l]{70}{wall}} 
                          & {\rotatebox[origin=l]{70}{fence}} 
                          & {\rotatebox[origin=l]{70}{pole}} 
                          & {\rotatebox[origin=l]{70}{light}} 
                          & {\rotatebox[origin=l]{70}{sign}} 
                          & {\rotatebox[origin=l]{70}{vegetation}} 
                          & {\rotatebox[origin=l]{70}{terrain}} 
                          & {\rotatebox[origin=l]{70}{sky}} 
                          & {\rotatebox[origin=l]{70}{person}} 
                          & {\rotatebox[origin=l]{70}{rider}} 
                          & {\rotatebox[origin=l]{70}{car}} 
                          & {\rotatebox[origin=l]{70}{truck}} 
                          & {\rotatebox[origin=l]{70}{bus}} 
                          & {\rotatebox[origin=l]{70}{train}} 
                          & {\rotatebox[origin=l]{70}{motocycle}} 
                          & {\rotatebox[origin=l]{70}{bicycle}} 
                          & {mIoU} \\ \midrule
                          
    {\multirow{11}{*}{  \begin{tabular}{@{}c@{}}ResNet101\cite{he2016deep} \\ 65.1\end{tabular}  }} 
      & AdaStruct\cite{tsai2018learning} & 86.5 & 25.9 & 79.8 & 22.1 & 20.0 & 23.6 & 33.1 & 21.8 & 81.8 & 25.9 & 75.9 & 57.3 & 26.2 & 76.3 & 29.8 & 32.1 & 7.2 & 29.5 & 32.5 & 41.4 \\
      & DCAN\cite{wu2018dcan} & 85.0 & 30.8 & 81.3 & 25.8 & 21.2 & 22.2 & 25.4 & 26.6 & 83.4 & 36.7 & 76.2 & 58.9 & 24.9 & 80.7 & 29.5 & 42.9 & 2.5 & 26.9 & 11.6 & 41.7 \\
      & DLOW\cite{gong2019dlow} & 87.1 & 33.5 & 80.5 & 24.5 & 13.2 & 29.8 & 29.5 & 26.6 & 82.6 & 26.7 & 81.8 & 55.9 & 25.3 & 78.0 & 33.5 & 38.7 & 0.0 & 22.9 & 34.5 & 42.3 \\
      & Cycada\cite{hoffman2018cycada} & 86.7 & 35.6 & 80.1 & 19.8 & 17.5 & 38.0 & 39.9 & 41.5 & 82.7 & 27.9 & 73.6 & 64.9 & 19 & 65.0 & 12.0 & 28.6 & 4.5 & 31.1 & 42.0 & 42.7 \\
      & CLAN\cite{luo2019taking} & 87.0 & 27.1 & 79.6 & 27.3 & 23.3 & 28.3 & 35.5 & 24.2 & 83.6 & 27.4 & 74.2 & 58.6 & 28.0 & 76.2 & 33.1 & 36.7 & 6.7 & 31.9 & 31.4 & 43.2 \\
      & ABStruct\cite{chang2019all} & 91.5 & 47.5 & 82.5 & 31.3 & 25.6 & 33.0 & 33.7 & 25.8 & 82.7 & 28.8 & 82.7 & 62.4 & 30.8 & 85.2 & 27.7 & 34.5 & 6.4 & 25.2 & 24.4 & 45.4 \\
      & AdvEnt\cite{vu2019advent} & 89.4 & 33.1 & 81.0 & 26.6 & 26.8 & 27.2 & 33.5 & 24.7 & 83.9 & 36.7 & 78.8 & 58.7 & 30.5 & 84.8 & 38.5 & 44.5 & 1.7 & 31.6 & 32.4 & 45.5 \\
      & BDL \cite{li2019bidirectional} & 91.0 & 44.7 & 84.2 & 34.6 & 27.6 & 30.2 & 36.0 & 36.0 & 85.0 & 43.6 & 83.0 & 58.6 & 31.6 & 83.3 & 35.3 & 49.7 & 3.3 & 28.8 & 35.6 & 48.5 \\
      & FDA & 90.0 & 40.5 & 79.4 & 25.3 & 26.7 & 30.6 & 31.9 & 29.3 & 79.4 & 28.8 & 76.5 & 56.4 & 27.5 & 81.7 & 27.7 & 45.1 & 17.0 & 23.8 & 29.6 & 44.6 \\
      & FDA-ENT & 90.8 & 42.7 & 80.8 & 28.1 & 26.6 & 31.8 & 32.8 & 29.1 & 81.6 & 31.2 & 76.2 & 56.9 & 27.7 & 82.8 & 25.3 & 44.1 & 15.3 & 21.1 & 30.2 & 45.0 \\
      & FDA-MBT & 92.5 & 53.3 & 82.4 & 26.5 & 27.6 & 36.4 & 40.6 & 38.9 & 82.3 & 39.8 & 78.0 & 62.6 & 34.4 & 84.9 & 34.1 & 53.1 & 16.9 & 27.7 & 46.4 & {\bf 50.45} \\ \midrule
      
      {\multirow{9}{*}{  \begin{tabular}{@{}c@{}}VGG16\cite{simonyan2014very} \\ 60.3\end{tabular}  }} 
      & CBST\cite{zou2018unsupervised} & 66.7 & 26.8 & 73.7 & 14.8 & 9.5 & 28.3 & 25.9 & 10.1 & 75.5 & 15.7 & 51.6 & 47.2 & 6.2 & 71.9 & 3.7 & 2.2 & 5.4 & 18.9 & 32.4 & 30.9 \\
      & SIBAN\cite{luo2019significance} & 83.4 & 13.0 & 77.8 & 20.4 & 17.5 & 24.6 & 22.8 & 9.6 & 81.3 & 29.6 & 77.3 & 42.7 & 10.9 & 76.0 & 22.8 & 17.9 & 5.7 & 14.2 & 2.0 & 34.2 \\
      & Cycada\cite{hoffman2018cycada} & 85.2 & 37.2 & 76.5 & 21.8 & 15.0 & 23.8 & 22.9 & 21.5 & 80.5 & 31.3 & 60.7 & 50.5 & 9.0 & 76.9 & 17.1 & 28.2 & 4.5 & 9.8 & 0 & 35.4 \\      
      & AdvEnt\cite{vu2019advent} & 86.9 & 28.7 & 78.7 & 28.5 & 25.2 & 17.1 & 20.3 & 10.9 & 80.0 & 26.4 & 70.2 & 47.1 & 8.4 & 81.5 & 26.0 & 17.2 & 18.9 & 11.7 & 1.6 & 36.1 \\
      & DCAN\cite{wu2018dcan} & 82.3 & 26.7 & 77.4 & 23.7 & 20.5 & 20.4 & 30.3 & 15.9 & 80.9 & 25.4 & 69.5 & 52.6 & 11.1 & 79.6 & 24.9 & 21.2 & 1.30 & 17.0 & 6.70 & 36.2 \\
      & CLAN\cite{luo2019taking} & 88.0 & 30.6 & 79.2 & 23.4 & 20.5 & 26.1 & 23.0 & 14.8 & 81.6 & 34.5 & 72.0 & 45.8 & 7.9 & 80.5 & 26.6 & 29.9 & 0.0 & 10.7 & 0.0 & 36.6 \\
      & LSD\cite{sankaranarayanan2018learning} & 88.0 & 30.5 & 78.6 & 25.2 & 23.5 & 16.7 & 23.5 & 11.6 & 78.7 & 27.2 & 71.9 & 51.3 & 19.5 & 80.4 & 19.8 & 18.3 & 0.9 & 20.8 & 18.4 & 37.1 \\
      & BDL \cite{li2019bidirectional} & 89.2 & 40.9 & 81.2 & 29.1 & 19.2 & 14.2 & 29.0 & 19.6 & 83.7 & 35.9 & 80.7 & 54.7 & 23.3 & 82.7 & 25.8 & 28.0 & 2.3 & 25.7 & 19.9 & 41.3\\
      & FDA-MBT & 86.1 & 35.1 & 80.6 & 30.8 & 20.4 & 27.5 & 30.0 & 26.0 & 82.1 & 30.3 & 73.6 & 52.5 & 21.7 & 81.7 & 24.0 & 30.5 & 29.9 & 14.6 & 24.0 & {\bf 42.2} \\ \bottomrule
\end{tabular}}
\end{center}
\vspace{-0.0cm}
\caption{{\it Quantitative Comparison on GTA5$\rightarrow$CityScapes.} The scores under each backbone represent the upper bound (train and test on the source domain). FDA: our method with a single scale; FDA-ENT: again single scale but with entropy regularization; FDA-MBT: FDA with multiple scales and Self-supervised Training. Note that our method consistently achieves better performance across different backbones.}
\vspace{-0.0cm}
\label{tab:benchmark_gta2city}
\end{table*}

{\bf SYNTHIA$\rightarrow$CityScapes}:

Following the evaluation protocol in \cite{li2019bidirectional}, we report the mIOU of our method on 16 classes using the VGG16 backbone, and on 13 classes using the ResNet101 backbone. Quantitative comparison is shown in Tab. \ref{tab:benchmark_syn2city}. Note again, our method achieves the top performance using different backbones and outperforms the seconder performer BDL \cite{li2019bidirectional} by 2.1\% and 3.9\%, respectively.

\setlength{\tabcolsep}{0.15em}
\begin{table*}
\begin{center}
\resizebox{0.9\textwidth}{!}{
\begin{tabular}{cccccccccccccccccccccc} \toprule
    {Backbone} & {Method} & {\rotatebox[origin=l]{70}{road}} 
                          & {\rotatebox[origin=l]{70}{sidewalk}} 
                          & {\rotatebox[origin=l]{70}{building}} 
                          & {\rotatebox[origin=l]{70}{wall}} 
                          & {\rotatebox[origin=l]{70}{fence}} 
                          & {\rotatebox[origin=l]{70}{pole}} 
                          & {\rotatebox[origin=l]{70}{light}} 
                          & {\rotatebox[origin=l]{70}{sign}} 
                          & {\rotatebox[origin=l]{70}{vegetation}} 
                          & {\rotatebox[origin=l]{70}{sky}} 
                          & {\rotatebox[origin=l]{70}{person}} 
                          & {\rotatebox[origin=l]{70}{rider}} 
                          & {\rotatebox[origin=l]{70}{car}} 
                          & {\rotatebox[origin=l]{70}{bus}} 
                          & {\rotatebox[origin=l]{70}{motocycle}} 
                          & {\rotatebox[origin=l]{70}{bicycle}} 
                          & {mIoU} \\ \midrule
      {\multirow{6}{*}{  \begin{tabular}{@{}c@{}}ResNet101\cite{he2016deep} \\ 71.7\end{tabular}  }} 
      & SIBAN\cite{luo2019significance} & 82.5 & 24.0 & 79.4 & - & - & - & 16.5 & 12.7 & 79.2 & 82.8 & 58.3 & 18.0 & 79.3 & 25.3 & 17.6 & 25.9 & 46.3 \\
      & CLAN\cite{luo2019taking} & 81.3 & 37.0 & 80.1 & - & - & - & 16.1 & 13.7 & 78.2 & 81.5 & 53.4 & 21.2 & 73.0 & 32.9 & 22.6 & 30.7 & 47.8 \\
      & ABStruct\cite{chang2019all} & 91.7 & 53.5 & 77.1 & - & - & - & 6.2 & 7.6 & 78.4 & 81.2 & 55.8 & 19.2 & 82.3 & 30.3 & 17.1 & 34.3 & 48.8 \\ 
      & AdvEnt\cite{vu2019advent} & 85.6 & 42.2 & 79.7 & - & - & - & 5.4 & 8.1 & 80.4 & 84.1 & 57.9 & 23.8 & 73.3 & 36.4 & 14.2 & 33.0 & 48.0 \\
      & BDL \cite{li2019bidirectional} & 86.0 & 46.7 & 80.3 & - & - & - & 14.1 & 11.6 & 79.2 & 81.3 & 54.1 & 27.9 & 73.7 & 42.2 & 25.7 & 45.3 & 51.4 \\
      & FDA-MBT & 79.3 & 35.0 & 73.2 & - & - & - & 19.9 & 24.0 & 61.7 & 82.6 & 61.4 & 31.1 & 83.9 & 40.8 & 38.4 & 51.1 & {\bf 52.5} \\ \midrule
      
      {\multirow{7}{*}{  \begin{tabular}{@{}c@{}}VGG16\cite{simonyan2014very} \\ 59.5\end{tabular}  }} 
      & AdvEnt\cite{vu2019advent} & 67.9 & 29.4 & 71.9 & 6.3 & 0.3 & 19.9 & 0.6 & 2.6 & 74.9 & 74.9 & 35.4 & 9.6 & 67.8 & 21.4 & 4.1 & 15.5 & 31.4 \\
      & DCAN\cite{wu2018dcan} & 79.9 & 30.4 & 70.8 & 1.6 & 0.6 & 22.3 & 6.7 & 23.0 & 76.9 & 73.9 & 41.9 & 16.7 & 61.7 & 11.5 & 10.3 & 38.6 & 35.4 \\
      & LSD\cite{sankaranarayanan2018learning} & 80.1 & 29.1 & 77.5 & 2.8 & 0.4 & 26.8 & 11.1 & 18.0 & 78.1 & 76.7 & 48.2 & 15.2 & 70.5 & 17.4 & 8.7 & 16.7 & 36.1 \\
      & ROAD\cite{chen2018road} & 77.7 & 30.0 & 77.5 & 9.6 & 0.3 & 25.8 & 10.3 & 15.6 & 77.6 & 79.8 & 44.5 & 16.6 & 67.8 & 14.5 & 7.0 & 23.8 & 36.2 \\
      & GIO-Ada\cite{chen2019learning} & 78.3 & 29.2 & 76.9 & 11.4 & 0.3 & 26.5 & 10.8 & 17.2 & 81.7 & 81.9 & 45.8 & 15.4 & 68.0 & 15.9 & 7.5 & 30.4 & 37.3 \\
      & BDL \cite{li2019bidirectional} & 72.0 & 30.3 & 74.5 & 0.1 & 0.3 & 24.6 & 10.2 & 25.2 & 80.5 & 80.0 & 54.7 & 23.2 & 72.7 & 24.0 & 7.5 & 44.9 & 39.0 \\
      & FDA-MBT & 84.2 & 35.1 & 78.0 & 6.1 & 0.44 & 27.0 & 8.5 & 22.1 & 77.2 & 79.6 & 55.5 & 19.9 & 74.8 & 24.9 & 14.3 & 40.7 & {\bf 40.5} \\ \bottomrule
\end{tabular}}
\end{center}
\caption{{\it Quantitative Comparison on SYNTHIA$\rightarrow$CityScapes.} Scores under each backbone represent the upper bound. For VGG, we evaluate on the 16 subclasses, and for ResNet101, 13 of the 16 classes are evaluated according to the evaluation protocol in the literature. Classes not evaluated are replaced by '-.' Our method consistently achieves better performance than the others across different backbones.}
\label{tab:benchmark_syn2city}
\end{table*}

\subsection{Qualitative Results}

\begin{figure*}[!ht]
\begin{center}
  \includegraphics[width=0.96\textwidth]{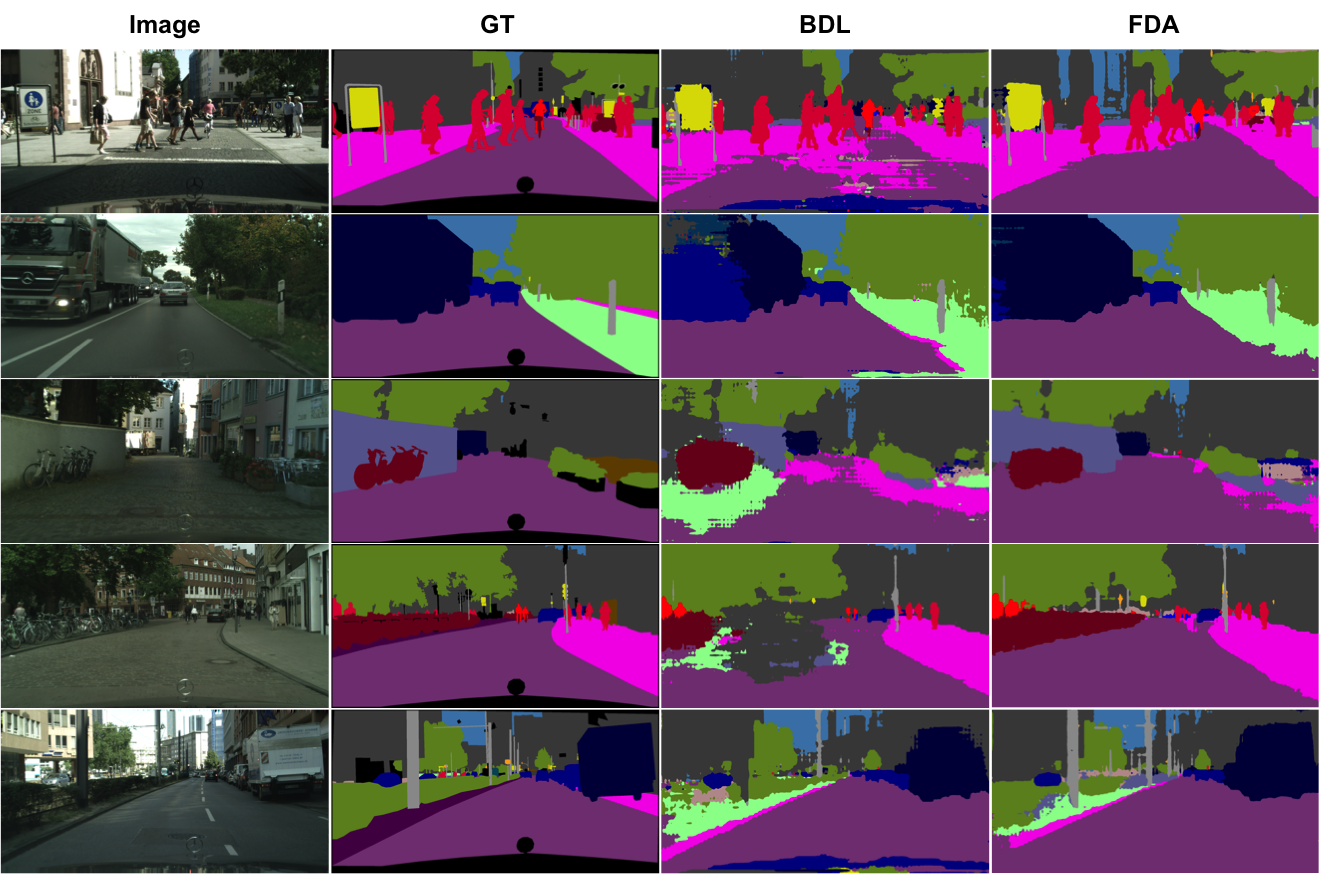}
\end{center}
\caption{ {\it Visual Comparison}. Left to right: Input image from CityScapes, ground-truth semantic segmentation, BDL \cite{li2019bidirectional}, FDA-MBT. Note that the predictions from FDA-MBT are generally smoother, e.g. the road in the first and fourth row, and the wall in the third row. Moreover, FDA-MBT achieves better performance on fine structures, e.g., the poles in the fifth row.}
\label{fig:compare-to-BDL}
\end{figure*}

We visually compare to the second performer BDL \cite{li2019bidirectional} who uses the same segmentation network backbone as ours. As we can see from Fig. \ref{fig:compare-to-BDL}, the predictions from our model appear much less noisy, like the road in the first row. Not only smoother, but our method can also maintain the fine structures, like the poles in the fifth row. Moreover, our method performs well on rare classes, for example, the truck in the second row, and the bicycles in the third and fourth rows. We accredit this to both the generalization ability of the single scale FDA, and the regularized SST by our Multi-band Transfer.

\section{Discussion}


We have proposed a simple method for domain alignment that does not require any learning, and can be easily integrated into a learning system that transforms unsupervised domain adaptation into semi-supervised learning. 
Some attention needs to be devoted to proper regularization of the loss function, for which we propose an entropic regularizer with anisotropic (Charbonnier) weighting. 
The self-referential problem in self-supervised training is addressed by the Multi-band Transfer scheme that requires no joint training of student networks with complicated model selection.

The results indicate that our method not only improves on the baseline, which was expected, but actually surpasses the current state of the art, which is considerably more involved, despite its simplicity. This suggests that some distributional misalignment due to low-level statistics, which has been known to wreak havoc with generalization across different domains, is quite simple to capture with a fast Fourier transform.
Moreover, the inverse Fourier transform of the spectrum of a real signal is guaranteed to be real, as one can easily show that the imaginary part is canceled given the skew-symmetry of the integrand; thus, images that are domain adapted using our method still reside in the real image space.

Robustness to nuisance variability affecting the image domain remains a difficult problem in machine learning, and we do not claim our method to be the final solution. However, we show that in some cases, it may not be necessary to learn what we already know, such as the fact that low-level statistics of the image can vary widely without affecting the semantics of the underlying scene. Such pre-processing could be an alternative to sophisticated architectures or laborious data augmentation. 
In the future, we would like to see applications of our method on other domain adaptation tasks.

\section*{Acknowledgements}

Research supported by ARO W911NF-17-1-0304 and ONR N00014-19-1-2066.

{\small
\bibliographystyle{ieee_fullname}
\bibliography{egbib}

\begin{thebibliography}{10}\itemsep=-1pt

\bibitem{achille2017critical}
Alessandro Achille, Matteo Rovere, and Stefano Soatto.
\newblock Critical learning periods in deep networks.
\newblock In {\em International Conference on Learning Representations}, 2019.

\bibitem{bruhn2005towards}
Andres Bruhn and Joachim Weickert.
\newblock Towards ultimate motion estimation: Combining highest accuracy with
  real-time performance.
\newblock In {\em Tenth IEEE International Conference on Computer Vision
  (ICCV'05) Volume 1}, volume~1, pages 749--755. IEEE, 2005.

\bibitem{cariucci2017autodial}
Fabio~Maria Cariucci, Lorenzo Porzi, Barbara Caputo, Elisa Ricci, and
  Samuel~Rota Bul{\`o}.
\newblock Autodial: Automatic domain alignment layers.
\newblock In {\em 2017 IEEE International Conference on Computer Vision
  (ICCV)}, pages 5077--5085. IEEE, 2017.

\bibitem{chang2019all}
Wei-Lun Chang, Hui-Po Wang, Wen-Hsiao Peng, and Wei-Chen Chiu.
\newblock All about structure: Adapting structural information across domains
  for boosting semantic segmentation.
\newblock In {\em Proceedings of the IEEE Conference on Computer Vision and
  Pattern Recognition}, pages 1900--1909, 2019.

\bibitem{chen2017deeplab}
Liang-Chieh Chen, George Papandreou, Iasonas Kokkinos, Kevin Murphy, and Alan~L
  Yuille.
\newblock Deeplab: Semantic image segmentation with deep convolutional nets,
  atrous convolution, and fully connected crfs.
\newblock {\em IEEE transactions on pattern analysis and machine intelligence},
  40(4):834--848, 2017.

\bibitem{chen2019learning}
Yuhua Chen, Wen Li, Xiaoran Chen, and Luc~Van Gool.
\newblock Learning semantic segmentation from synthetic data: A geometrically
  guided input-output adaptation approach.
\newblock In {\em Proceedings of the IEEE Conference on Computer Vision and
  Pattern Recognition}, pages 1841--1850, 2019.

\bibitem{chen2018road}
Yuhua Chen, Wen Li, and Luc Van~Gool.
\newblock Road: Reality oriented adaptation for semantic segmentation of urban
  scenes.
\newblock In {\em Proceedings of the IEEE Conference on Computer Vision and
  Pattern Recognition}, pages 7892--7901, 2018.

\bibitem{choi2018stargan}
Yunjey Choi, Minje Choi, Munyoung Kim, Jung-Woo Ha, Sunghun Kim, and Jaegul
  Choo.
\newblock Stargan: Unified generative adversarial networks for multi-domain
  image-to-image translation.
\newblock In {\em Proceedings of the IEEE Conference on Computer Vision and
  Pattern Recognition}, pages 8789--8797, 2018.

\bibitem{cordts2016cityscapes}
Marius Cordts, Mohamed Omran, Sebastian Ramos, Timo Rehfeld, Markus Enzweiler,
  Rodrigo Benenson, Uwe Franke, Stefan Roth, and Bernt Schiele.
\newblock The cityscapes dataset for semantic urban scene understanding.
\newblock In {\em Proceedings of the IEEE conference on computer vision and
  pattern recognition}, pages 3213--3223, 2016.

\bibitem{csurka2017domain}
Gabriela Csurka.
\newblock A comprehensive survey on domain adaptation for visual applications.
\newblock In {\em Domain adaptation in computer vision applications}, pages
  1--35. Springer, 2017.

\bibitem{everingham2015pascal}
Mark Everingham, SM~Ali Eslami, Luc Van~Gool, Christopher~KI Williams, John
  Winn, and Andrew Zisserman.
\newblock The pascal visual object classes challenge: A retrospective.
\newblock {\em International journal of computer vision}, 111(1):98--136, 2015.

\bibitem{french2017self}
Geoffrey French, Michal Mackiewicz, and Mark Fisher.
\newblock Self-ensembling for visual domain adaptation.
\newblock In {\em International Conference on Learning Representations},
  number~6, 2018.

\bibitem{frigo1998fftw}
Matteo Frigo and Steven~G Johnson.
\newblock Fftw: An adaptive software architecture for the fft.
\newblock In {\em Proceedings of the 1998 IEEE International Conference on
  Acoustics, Speech and Signal Processing, ICASSP'98 (Cat. No. 98CH36181)},
  volume~3, pages 1381--1384. IEEE, 1998.

\bibitem{ganin2014unsupervised}
Yaroslav Ganin and Victor Lempitsky.
\newblock Unsupervised domain adaptation by backpropagation.
\newblock In {\em International Conference on Machine Learning}, pages
  1180--1189, 2015.

\bibitem{geng2011daml}
Bo Geng, Dacheng Tao, and Chao Xu.
\newblock Daml: Domain adaptation metric learning.
\newblock {\em IEEE Transactions on Image Processing}, 20(10):2980--2989, 2011.

\bibitem{ghifary2016deep}
Muhammad Ghifary, W~Bastiaan Kleijn, Mengjie Zhang, David Balduzzi, and Wen Li.
\newblock Deep reconstruction-classification networks for unsupervised domain
  adaptation.
\newblock In {\em European Conference on Computer Vision}, pages 597--613.
  Springer, 2016.

\bibitem{gong2019dlow}
Rui Gong, Wen Li, Yuhua Chen, and Luc~Van Gool.
\newblock Dlow: Domain flow for adaptation and generalization.
\newblock In {\em Proceedings of the IEEE Conference on Computer Vision and
  Pattern Recognition}, pages 2477--2486, 2019.

\bibitem{he2016deep}
Kaiming He, Xiangyu Zhang, Shaoqing Ren, and Jian Sun.
\newblock Deep residual learning for image recognition.
\newblock In {\em Proceedings of the IEEE conference on computer vision and
  pattern recognition}, pages 770--778, 2016.

\bibitem{hoffman2018cycada}
Judy Hoffman, Eric Tzeng, Taesung Park, Jun-Yan Zhu, Phillip Isola, Kate
  Saenko, Alexei Efros, and Trevor Darrell.
\newblock Cycada: Cycle-consistent adversarial domain adaptation.
\newblock In {\em Proceedings of the 35th International Conference on Machine
  Learning}, 2018.

\bibitem{hoffman2016fcns}
Judy Hoffman, Dequan Wang, Fisher Yu, and Trevor Darrell.
\newblock Fcns in the wild: Pixel-level adversarial and constraint-based
  adaptation.
\newblock {\em arXiv preprint arXiv:1612.02649}, 2016.

\bibitem{kumar2018co}
Abhishek Kumar, Prasanna Sattigeri, Kahini Wadhawan, Leonid Karlinsky, Rogerio
  Feris, Bill Freeman, and Gregory Wornell.
\newblock Co-regularized alignment for unsupervised domain adaptation.
\newblock In {\em Advances in Neural Information Processing Systems}, pages
  9345--9356, 2018.

\bibitem{laine2016temporal}
Samuli Laine and Timo Aila.
\newblock Temporal ensembling for semi-supervised learning.
\newblock {\em arXiv preprint arXiv:1610.02242}, 2016.

\bibitem{li2019bidirectional}
Yunsheng Li, Lu Yuan, and Nuno Vasconcelos.
\newblock Bidirectional learning for domain adaptation of semantic
  segmentation.
\newblock In {\em Proceedings of the IEEE Conference on Computer Vision and
  Pattern Recognition}, pages 6936--6945, 2019.

\bibitem{lin2014microsoft}
Tsung-Yi Lin, Michael Maire, Serge Belongie, James Hays, Pietro Perona, Deva
  Ramanan, Piotr Doll{\'a}r, and C~Lawrence Zitnick.
\newblock Microsoft coco: Common objects in context.
\newblock In {\em European conference on computer vision}, pages 740--755.
  Springer, 2014.

\bibitem{liu2017unsupervised}
Ming-Yu Liu, Thomas Breuel, and Jan Kautz.
\newblock Unsupervised image-to-image translation networks.
\newblock In {\em Advances in neural information processing systems}, pages
  700--708, 2017.

\bibitem{long2015fully}
Jonathan Long, Evan Shelhamer, and Trevor Darrell.
\newblock Fully convolutional networks for semantic segmentation.
\newblock In {\em Proceedings of the IEEE conference on computer vision and
  pattern recognition}, pages 3431--3440, 2015.

\bibitem{long2015learning}
Mingsheng Long, Yue Cao, Jianmin Wang, and Michael Jordan.
\newblock Learning transferable features with deep adaptation networks.
\newblock In {\em International Conference on Machine Learning}, pages 97--105,
  2015.

\bibitem{luo2019significance}
Yawei Luo, Ping Liu, Tao Guan, Junqing Yu, and Yi Yang.
\newblock Significance-aware information bottleneck for domain adaptive
  semantic segmentation.
\newblock In {\em Proceedings of the IEEE International Conference on Computer
  Vision}, pages 6778--6787, 2019.

\bibitem{luo2019taking}
Yawei Luo, Liang Zheng, Tao Guan, Junqing Yu, and Yi Yang.
\newblock Taking a closer look at domain shift: Category-level adversaries for
  semantics consistent domain adaptation.
\newblock In {\em Proceedings of the IEEE Conference on Computer Vision and
  Pattern Recognition}, pages 2507--2516, 2019.

\bibitem{mancini2018boosting}
Massimiliano Mancini, Lorenzo Porzi, Samuel Rota~Bul{\`o}, Barbara Caputo, and
  Elisa Ricci.
\newblock Boosting domain adaptation by discovering latent domains.
\newblock In {\em Proceedings of the IEEE Conference on Computer Vision and
  Pattern Recognition}, pages 3771--3780, 2018.

\bibitem{motiian2017unified}
Saeid Motiian, Marco Piccirilli, Donald~A Adjeroh, and Gianfranco Doretto.
\newblock Unified deep supervised domain adaptation and generalization.
\newblock In {\em Proceedings of the IEEE International Conference on Computer
  Vision}, pages 5715--5725, 2017.

\bibitem{patel2015visual}
Vishal~M Patel, Raghuraman Gopalan, Ruonan Li, and Rama Chellappa.
\newblock Visual domain adaptation: A survey of recent advances.
\newblock {\em IEEE signal processing magazine}, 32(3):53--69, 2015.

\bibitem{richter2016playing}
Stephan~R Richter, Vibhav Vineet, Stefan Roth, and Vladlen Koltun.
\newblock Playing for data: Ground truth from computer games.
\newblock In {\em European conference on computer vision}, pages 102--118.
  Springer, 2016.

\bibitem{ros2016synthia}
German Ros, Laura Sellart, Joanna Materzynska, David Vazquez, and Antonio~M
  Lopez.
\newblock The synthia dataset: A large collection of synthetic images for
  semantic segmentation of urban scenes.
\newblock In {\em Proceedings of the IEEE conference on computer vision and
  pattern recognition}, pages 3234--3243, 2016.

\bibitem{rosenberg2005semi}
Chuck Rosenberg, Martial Hebert, and Henry Schneiderman.
\newblock Semi-supervised self-training of object detection models.
\newblock In {\em Proceedings of the Seventh IEEE Workshops on Application of
  Computer Vision (WACV/MOTION'05)-Volume 1-Volume 01}, pages 29--36, 2005.

\bibitem{saito2017asymmetric}
Kuniaki Saito, Yoshitaka Ushiku, and Tatsuya Harada.
\newblock Asymmetric tri-training for unsupervised domain adaptation.
\newblock In {\em Proceedings of the 34th International Conference on Machine
  Learning-Volume 70}, pages 2988--2997. JMLR. org, 2017.

\bibitem{sankaranarayanan2018learning}
Swami Sankaranarayanan, Yogesh Balaji, Arpit Jain, Ser Nam~Lim, and Rama
  Chellappa.
\newblock Learning from synthetic data: Addressing domain shift for semantic
  segmentation.
\newblock In {\em Proceedings of the IEEE Conference on Computer Vision and
  Pattern Recognition}, pages 3752--3761, 2018.

\bibitem{sener2016learning}
Ozan Sener, Hyun~Oh Song, Ashutosh Saxena, and Silvio Savarese.
\newblock Learning transferrable representations for unsupervised domain
  adaptation.
\newblock In {\em Advances in Neural Information Processing Systems}, pages
  2110--2118, 2016.

\bibitem{shu2018dirt}
Rui Shu, Hung~H Bui, Hirokazu Narui, and Stefano Ermon.
\newblock A dirt-t approach to unsupervised domain adaptation.
\newblock In {\em Proc. 6th International Conference on Learning
  Representations}, 2018.

\bibitem{simonyan2014very}
Karen Simonyan and Andrew Zisserman.
\newblock Very deep convolutional networks for large-scale image recognition.
\newblock {\em arXiv preprint arXiv:1409.1556}, 2014.

\bibitem{sun2019deep}
Ke Sun, Bin Xiao, Dong Liu, and Jingdong Wang.
\newblock Deep high-resolution representation learning for human pose
  estimation.
\newblock In {\em Proceedings of the IEEE Conference on Computer Vision and
  Pattern Recognition}, pages 5693--5703, 2019.

\bibitem{tarvainen2017mean}
Antti Tarvainen and Harri Valpola.
\newblock Mean teachers are better role models: Weight-averaged consistency
  targets improve semi-supervised deep learning results.
\newblock In {\em Advances in neural information processing systems}, pages
  1195--1204, 2017.

\bibitem{tsai2018learning}
Yi-Hsuan Tsai, Wei-Chih Hung, Samuel Schulter, Kihyuk Sohn, Ming-Hsuan Yang,
  and Manmohan Chandraker.
\newblock Learning to adapt structured output space for semantic segmentation.
\newblock In {\em Proceedings of the IEEE Conference on Computer Vision and
  Pattern Recognition}, pages 7472--7481, 2018.

\bibitem{tzeng2017adversarial}
Eric Tzeng, Judy Hoffman, Kate Saenko, and Trevor Darrell.
\newblock Adversarial discriminative domain adaptation.
\newblock In {\em Proceedings of the IEEE Conference on Computer Vision and
  Pattern Recognition}, pages 7167--7176, 2017.

\bibitem{vu2019advent}
Tuan-Hung Vu, Himalaya Jain, Maxime Bucher, Matthieu Cord, and Patrick
  P{\'e}rez.
\newblock Advent: Adversarial entropy minimization for domain adaptation in
  semantic segmentation.
\newblock In {\em Proceedings of the IEEE Conference on Computer Vision and
  Pattern Recognition}, pages 2517--2526, 2019.

\bibitem{wang2018deep}
Mei Wang and Weihong Deng.
\newblock Deep visual domain adaptation: A survey.
\newblock {\em Neurocomputing}, 312:135--153, 2018.

\bibitem{wu2018dcan}
Zuxuan Wu, Xintong Han, Yen-Liang Lin, Mustafa Gokhan~Uzunbas, Tom Goldstein,
  Ser Nam~Lim, and Larry~S Davis.
\newblock Dcan: Dual channel-wise alignment networks for unsupervised scene
  adaptation.
\newblock In {\em Proceedings of the European Conference on Computer Vision
  (ECCV)}, pages 518--534, 2018.

\bibitem{yang2020phase}
Yanchao Yang, Dong Lao, Ganesh Sundaramoorthi, and Stefano Soatto.
\newblock Phase consistent ecological domain adaptation.
\newblock In {\em Proceedings of the IEEE Conference on Computer Vision and
  Pattern Recognition}, 2020.

\bibitem{yang2018conditional}
Yanchao Yang and Stefano Soatto.
\newblock Conditional prior networks for optical flow.
\newblock In {\em Proceedings of the European Conference on Computer Vision
  (ECCV)}, pages 271--287, 2018.

\bibitem{yi2017dualgan}
Zili Yi, Hao Zhang, Ping Tan, and Minglun Gong.
\newblock Dualgan: Unsupervised dual learning for image-to-image translation.
\newblock In {\em Proceedings of the IEEE international conference on computer
  vision}, pages 2849--2857, 2017.

\bibitem{yu2015multi}
Fisher Yu and Vladlen Koltun.
\newblock Multi-scale context aggregation by dilated convolutions.
\newblock {\em arXiv preprint arXiv:1511.07122}, 2015.

\bibitem{zellinger2017central}
Werner Zellinger, Thomas Grubinger, Edwin Lughofer, Thomas Natschl{\"a}ger, and
  Susanne Saminger-Platz.
\newblock Central moment discrepancy (cmd) for domain-invariant representation
  learning.
\newblock {\em arXiv preprint arXiv:1702.08811}, 2017.

\bibitem{zhang2017curriculum}
Yang Zhang, Philip David, and Boqing Gong.
\newblock Curriculum domain adaptation for semantic segmentation of urban
  scenes.
\newblock In {\em Proceedings of the IEEE International Conference on Computer
  Vision}, pages 2020--2030, 2017.

\bibitem{zhao2017pyramid}
Hengshuang Zhao, Jianping Shi, Xiaojuan Qi, Xiaogang Wang, and Jiaya Jia.
\newblock Pyramid scene parsing network.
\newblock In {\em Proceedings of the IEEE conference on computer vision and
  pattern recognition}, pages 2881--2890, 2017.

\bibitem{zhu2017unpaired}
Jun-Yan Zhu, Taesung Park, Phillip Isola, and Alexei~A Efros.
\newblock Unpaired image-to-image translation using cycle-consistent
  adversarial networks.
\newblock In {\em Proceedings of the IEEE international conference on computer
  vision}, pages 2223--2232, 2017.

\bibitem{zou2018unsupervised}
Yang Zou, Zhiding Yu, BVK Vijaya~Kumar, and Jinsong Wang.
\newblock Unsupervised domain adaptation for semantic segmentation via
  class-balanced self-training.
\newblock In {\em Proceedings of the European Conference on Computer Vision
  (ECCV)}, pages 289--305, 2018.

\end{thebibliography}
}

\end{document}